\documentclass[twoside, print]{ieeecolor_arxiv}

\usepackage{tmi}
\usepackage{cite}

\usepackage{diagbox}
\usepackage{cite}
\usepackage{amsmath,amssymb,amsfonts,bm}
\usepackage{booktabs}
\usepackage{graphicx,epstopdf}
\usepackage{textcomp}
\usepackage{caption}
\usepackage{url}
\usepackage{multirow}
\usepackage{float}
\usepackage{tabularx,threeparttable}
\usepackage{makecell}
\usepackage{xcolor}
\usepackage[normalem]{ulem} 
\usepackage{tabulary}
\usepackage[colorlinks=true,citecolor=blue,urlcolor=blue]{hyperref}
\usepackage[]{color}
\usepackage{hhline}
\usepackage{array}
\usepackage{amsfonts}
\usepackage{amsmath}
\usepackage{mathabx}
\usepackage{fancyhdr}
\usepackage{soul}
\usepackage{makecell}
\usepackage{bbm}
\usepackage{amssymb}
\usepackage{pifont}
\newcommand{\xmark}{\ding{55}}%
\usepackage[noend]{algpseudocode}
\usepackage{algorithm}
\usepackage{flexisym}
\usepackage{eqparbox}
\usepackage{breqn}
\usepackage[]{silence}
\renewcommand{\algorithmiccomment}[1]{\textsf\bgroup\tiny//~#1\egroup}
\def\BibTeX{{\rm B\kern-.05em{\sc i\kern-.025em b}\kern-.08em
    T\kern-.1667em\lower.7ex\hbox{E}\kern-.125emX}}
\begin{document}
\title{Temporally Constrained Neural Networks (TCNN): A framework for semi-supervised video semantic segmentation}
%
\author{Deepak Alapatt, Pietro Mascagni, Armine Vardazaryan, Alain Garcia, Nariaki Okamoto, Didier Mutter, Jacques Marescaux, Guido Costamagna, Bernard Dallemagne and Nicolas Padoy
\thanks{
 This work was partially supported by French State Funds managed by the “Agence Nationale de la Recherche (ANR)” through the “Investissements d’Avenir” (Investments for the Future) Program under Grant ANR-10-IAHU-02 (IHU-Strasbourg) and through the National AI Chair program under Grant ANR-20-CHIA-0029-01 (Chair AI4ORSafety). This work was granted access to the HPC resources of IDRIS under the allocation 2021-AD011011640R1 made by GENCI.}
\thanks{Deepak Alapatt, and Nicolas Padoy are affiliated with ICube, University of Strasbourg, CNRS, France
(email: \{alapatt, npadoy\}@unistra.fr)
}
\thanks{Pietro Mascagni and Guido Costamagna are affiliated with Fondazione Policlinico Universitario A. Gemelli IRCCS, Rome, Italy}
\thanks{Pietro Mascagni, Armine Vardazaryan, Alain Garcia, Didier Mutter and Nicolas Padoy are affiliated with IHU-Strasbourg, Institute of Image-Guided Surgery, Strasbourg, France}
\thanks{Nariaki Okamato, Jacques Marescaux and Bernard Dallemagne are affiliated with Institute for Research against Digestive Cancer (IRCAD), Strasbourg, France}
}

\maketitle

\begin{abstract}
A major obstacle to building models for effective semantic segmentation, and particularly video semantic segmentation, is a lack of large and well annotated datasets. This bottleneck is particularly prohibitive in highly specialized and regulated fields such as medicine and surgery, where video semantic segmentation could have important applications but data and expert annotations are scarce. In these settings, temporal clues and anatomical constraints could be leveraged during training to improve performance. Here, we present Temporally Constrained Neural Networks (TCNN), a semi-supervised framework used for video semantic segmentation of surgical videos. In this work, we show that autoencoder networks can be used to efficiently provide both spatial and temporal supervisory signals to train deep learning models. We test our method on a newly introduced video dataset of laparoscopic cholecystectomy procedures, Endoscapes, and an adaptation of a public dataset of cataract surgeries, CaDIS. We demonstrate that lower-dimensional representations of predicted masks can be leveraged to provide a consistent improvement on both sparsely labeled datasets with no additional computational cost at inference time. Further, the TCNN framework is model-agnostic and can be used in conjunction with other model design choices with minimal additional complexity. 
\end{abstract}


\section{Introduction}
\label{sec:introduction}
    \IEEEPARstart{I}{mage-guided} interventions are becoming the mainstay of surgery. The analysis of the videos natively guiding such minimally invasive interventions holds great potential to study and assist surgical procedures \cite{mascagni2021or}. In this context, video analysis for surgical scene understanding is a key enabler towards context-aware computer-assisted interventions \cite{vercauteren2019cai4cai}, post-operative analysis of surgical procedures for documentation \cite{mascagni2021computer}, research and education \cite{ward2021surgical}, and eventually, robot-assisted surgery \cite{kennedy2020computer}. The task of image semantic segmentation presents a straightforward path to scene understanding through the categorization of every pixel of an image based on the structure it constitutes. However, a major limitation is the significant annotation effort and opportunity cost that annotating segmentation datasets entails for medical professionals, particularly when annotating fine-grained structures with high surgical semantics. A natural consequence is a lack of large and well-annotated surgical segmentation datasets compared to the computer vision community at large. As a result, there has been an increasing focus on optimally using the limited and precious datasets that are available.


Overall, the extension of image semantic segmentation methods to video-based applications, or video semantic segmentation, remains a challenging and open problem. One shortcoming of applying methods designed for image semantic segmentation to videos is that they fail to leverage the rich and abundant relationships between frames of videos. Moreover, proposed approaches for video semantic segmentation are often complex and computationally demanding. 

Further, traditional approaches treat semantic segmentation as a purely local problem, often training models using loss functions that penalize errors only at a pixel level. This approach has proven effective when working with large and well labeled datasets. However, intrinsic regularities in the data, such as the spatial relationship between anatomical structures in surgical videos, could be leveraged in the output space to boost performance even on smaller datasets.

To address these challenges, we propose a framework for semi-supervised video semantic segmentation, Temporally Constrained Neural Networks (TCNN). To validate our approach, we benchmark our method on two challenging surgical video datasets for both tool and anatomy segmentation. We believe this is the first work explicitly designed on the video semantic segmentation of anatomical classes in surgical videos.
  
Our contributions can be summarized as follows: 
\begin{itemize}
  \item We generate a new dataset for the segmentation of fine-grained anatomical structures in surgical videos.
  \item We demonstrate that autoencoder networks can be used to regularize models on highly variable and dynamic surgical videos.
  \item We propose a new framework for semi-supervised video semantic segmentation that is independent of model design choice and incurs no additional inference computational cost.
  \item We present results and analysis of our method in the context of two challenging surgical video datasets. We pick two datasets representing distinct surgical procedures, laparoscopic cholecystectomy and microscopic cataract surgery, that differ greatly in appearance and dynamism in order to illustrate the generalizability of our approach.
\end{itemize}

\begin{figure*}[!t]
\centerline{\includegraphics[clip,trim=0.0cm 0.6cm 0cm 2.8cm, width=17cm]{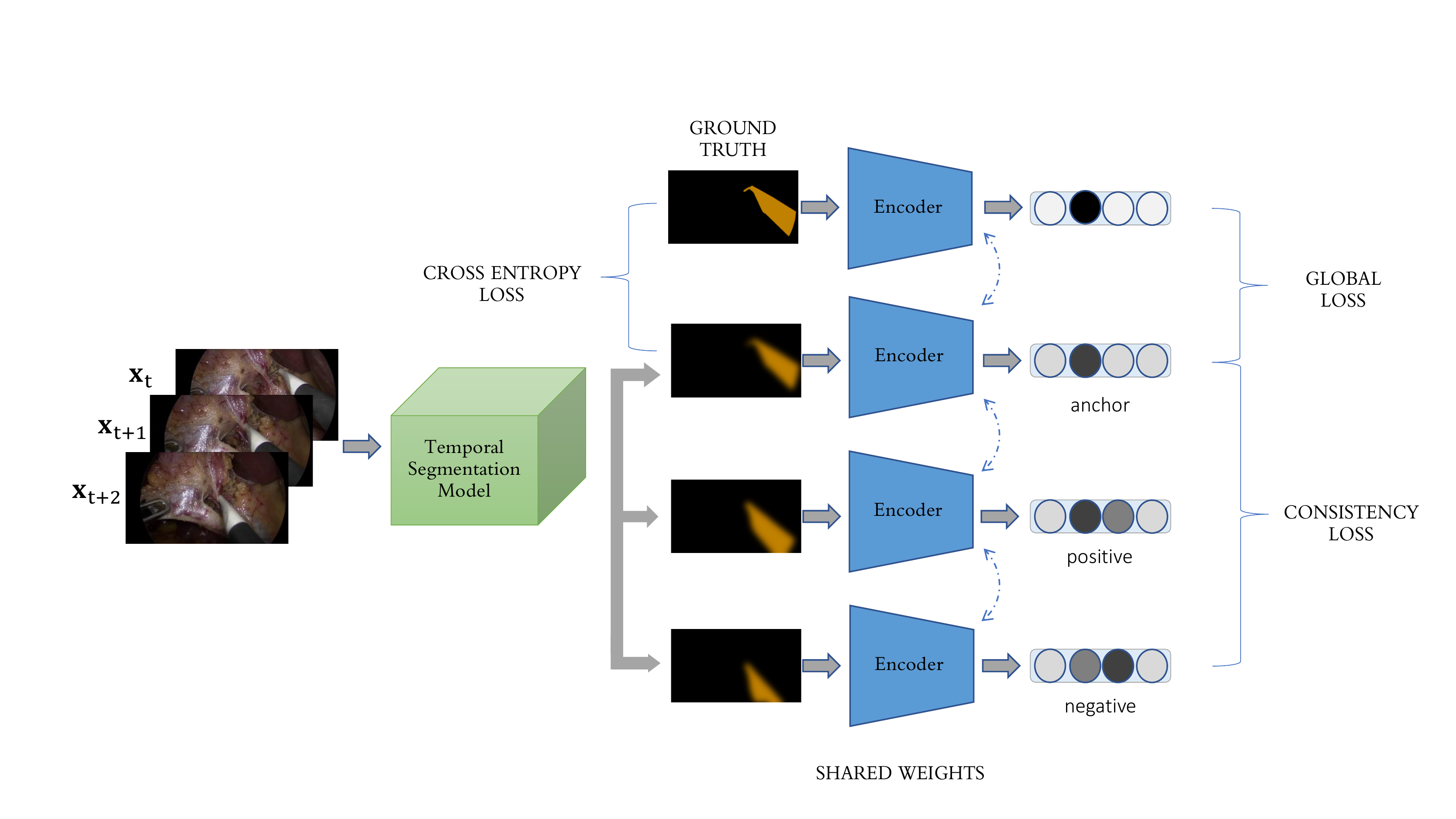}}
\caption{An overview of the proposed TCNN framework.}
\label{fig1}
\end{figure*}  
  
\section{Context}

\subsection{Image and Video Semantic Segmentation}
Image semantic segmentation is an active research field that has seen significant progress since the pioneering work applying fully convolutional networks for the task \cite{long2015fully}. Subsequent methods have focused on high quality\cite{hq1}\cite{hq2}\cite{hq3}\cite{hq4}\cite{hq5}\cite{hq6} and/or efficient \cite{efficient1}\cite{efficient2}\cite{efficient3}\cite{efficient4} design choices. 
More recently, the design of models for video semantic segmentation has received increasing attention. These methods largely fall into two categories: (1) approaches that reuse high-level features inferred from key frames for inference at surrounding frames \cite{keyframe1}\cite{keyframe2}\cite{keyframe3}\cite{keyframe4}; (2) approaches that run deep neural networks on every frame and use additional network layers to aggregate temporal information\cite{temporallayer1}\cite{temporallayer2}\cite{temporallayer3} \cite{temporallayer4}\cite{noisylstm}
. Due to a lack of densely labeled datasets, most methods in both categories use unlabeled frames to incorporate temporal context in their model training in a semi-supervised manner. However, the former category of methods primarily does so to minimize inference time, often at the cost of prediction performance\cite{keyframe3}\cite{keyframe4}, while the latter category focuses on using temporal information to improve predictions.  Our work falls into the second group but unlike most of the presented methods, our work is not restricted to a particular model design choice but rather presents a model-agnostic framework for spatio-temporal regularization of deep neural networks. For video semantic segmentation, in general, the use of optical flow appears to be the dominant strategy. However, flow-based methods often incur significant additional computational costs and are prone to failure when dealing with large motions, non-textured regions, and occlusions. These conditions are common in surgical videos and are further exacerbated by the highly similar appearance of anatomical structures. In the context of video semantic segmentation on surgical data, the topic remains largely unexplored barring a few works that have utilized optical flow for segmenting surgical instruments \cite{surgeryinstrumentflow1}\cite{surgeryinstrumentflow2}. To the best of our knowledge, no previously published work has been designed explicitly for the video semantic segmentation of anatomical classes. This may be because of the limited publicly available datasets or the additional complexity of the task.

\subsection{Autoencoders and Semantic Segmentation}
Most methods designed for semantic segmentation rely on loss functions such as cross entropy or dice loss to supervise the training process by penalizing errors made at a pixel level. As stated in \cite{acnn}, this treatment is suboptimal, particularly in constrained settings such as surgical video segmentation, since non-local features such as smoothness and topology are not explicitly taken into consideration. Moreover, learning constraints in the shape, position, and interaction between classes implicitly can be challenging without sizeable and representative datasets, as is common in the medical space.
Our work closely relates to a group of recently proposed methods that address these limitations using autoencoder networks. \cite{ravishankar} first introduced a cascaded architecture of a segmentation network followed by a denoising autoencoder, which was trained to map segmentation masks into a shape space before reconstructing them back. By incorporating an additional loss term that ensures that the encoded representations for the ground truth mask and the predicted mask lie close by in the shape space, they could train the network to learn non-local regularities in the output space such as shape. Along the same lines, \cite{acnn} use a similar cascaded design except using a frozen encoder to prevent over-regularization. \cite{postdae1}\cite{postdae2} utilize a similar mask denoising autoencoder as a post-processing step to "filter out" irregularities in the output space such as isolated blobs. Different from these methods, we demonstrate that an autoencoder trained to map segmentation masks to a lower-dimensional shape space could be further utilized to provide temporal supervision using the surrounding unlabeled frames in a sequence. In addition, we test the proposed framework on surgical videos, much more dynamic and unconstrained than the volumetric radiological imaging used to evaluate the methods described in this subsection.

\subsection{Video Semantic Segmentation Datasets}

As previously stated, generating large and representative datasets for video semantic segmentation can be extremely challenging. Most available datasets either limit the number of sequences being annotated ($\leq$10)\cite{camvid} or provide sparse annotations ($>$1Hz)\cite{cityscapes}. While a few publicly available video semantic segmentation datasets do exist in the computer vision community, similar datasets in the surgical domain are lacking. The few video datasets that do exist provide only labels for instrument classes \cite{robustmis2017}\cite{cata7} or label very few sequences(19)\cite{2018roboscene}. Recently, CholecSeg8k\cite{cholecseg8k} was introduced, providing anatomy and instrument labels for 101 selected short sequences ($\sim$3 seconds) from 17 publicly available laparoscopic cholecystectomy (LC) procedure recordings \cite{endonet}. In this paper, we generate and use Endoscapes, a new dataset consisting of 201 endoscopic video recordings of LC procedures annotated with segmentation masks of fine-grained anatomical structures. Compared to CholecSeg8k, Endoscapes comprises many more surgical procedures (201 vs 17), annotated classes (29 vs 13) and provides sparse annotations, every 30 seconds, rather than densely labeled short sequences. Additionally, we adapt a public image semantic segmentation dataset for cataract surgery, CaDIS \cite{cadis}, to generate temporal sequences. We hope that this work will promote research into video semantic segmentation for surgical scenes, which comes with a unique set of characteristics and challenges that warrant independent and further study; for example:

\begin{itemize}
    \item Surgical images may present a number of unique confounding factors such as smoke, bleeding, unstable camera motion which may not be as common in natural images.
    \item There may be an extremely high inter-class similarity between certain anatomical structures.
    \item In contrast to most video datasets, surgical datasets typically depict an evolving scene rather than a changing one, i.e. the same environment and structures are usually in focus.
\end{itemize}

\section{Endoscapes}
Cholecystectomy is a very common abdominal surgical procedure almost ubiquitously performed with a laparoscopic approach, hence guided by an endoscopic video. Deep learning models for LC video analysis have been developed with the aim of assisting surgeons during interventions \cite{mascagni2021artificial}\cite{ madani2021artificial}, improving staff awareness and readiness, and facilitating postoperative documentation and research \cite{mascagni2021computer}. However, datasets and models for video semantic segmentation of LC are lacking. Recognizing fine-grained hepatocystic anatomy through semantic segmentation could help surgeons better assess the critical view of safety (CVS), a universally recommended technique consisting in well exposing anatomical landmarks to prevent bile duct injuries \cite{brunt2020safe}. Additionally, segmentation masks of hepatocystic structures could be leveraged by deep learning models for automatic assessment of CVS \cite{mascagni2021artificial} and surgical action recognition \cite{nwoye2020recognition} to improve their performance. We believe that generating a dataset for video semantic segmentation of hepatocystic anatomy will promote surgical data science research and accelerate the development of applications for surgical safety.
To generate a representative dataset, consecutive endoscopic videos of  LC performed at Nouvel Hôpital Civil (Strasbourg, France) were collected. Non-endoscopic, i.e., out-of-body, video frames were blacked-out to comply with European data protection regulations. A frame every 30 seconds was sampled from the portion of the endoscopic video showing the hepatocystic anatomy being dissected, the most critical phase of the surgical procedure, and when surgeons should achieve the CVS. Such unselected and regularly spaced video frames were manually annotated with pixel-wise semantic annotations of anatomical and surgical instances, such as the cystic artery and the dissection of the hepatocystic triangle, respectively. Annotations were performed in double by specifically trained computer scientists and surgeons. For reproducibility and transparency, an in-depth description of the annotation process can be found in \cite{annotationprotocol}. 
Overall, 1933 regularly spaced video frames from 201 LC videos were annotated with segmentation mask for 29 classes. A more detailed description of the annotated classes and their representation in Endoscapes is presented in Table 1.

\section{Temporally Constrained Neural Networks }

In this section, we present a framework for semi-supervised semantic segmentation, which we term Temporally Constrained Neural Networks (TCNN). Our work is inspired by Anatomically Constrained Neural Networks\cite{acnn}, a recently proposed framework utilizing autoencoder\cite{denoisingautoencoders} networks to penalize global features in addition to local penalties such as cross entropy which are a standard for semantic segmentation. We extend this work to the much more dynamic and variable domain of surgical videos and further utilize the autoencoder network to supervise network training using the global and temporal context of the prediction. A brief overview of our framework is presented in Fig 1.

\begin{figure}[]
\centerline{\includegraphics[clip,trim=5.5cm 2cm 10cm 3cm, width=8.5cm]{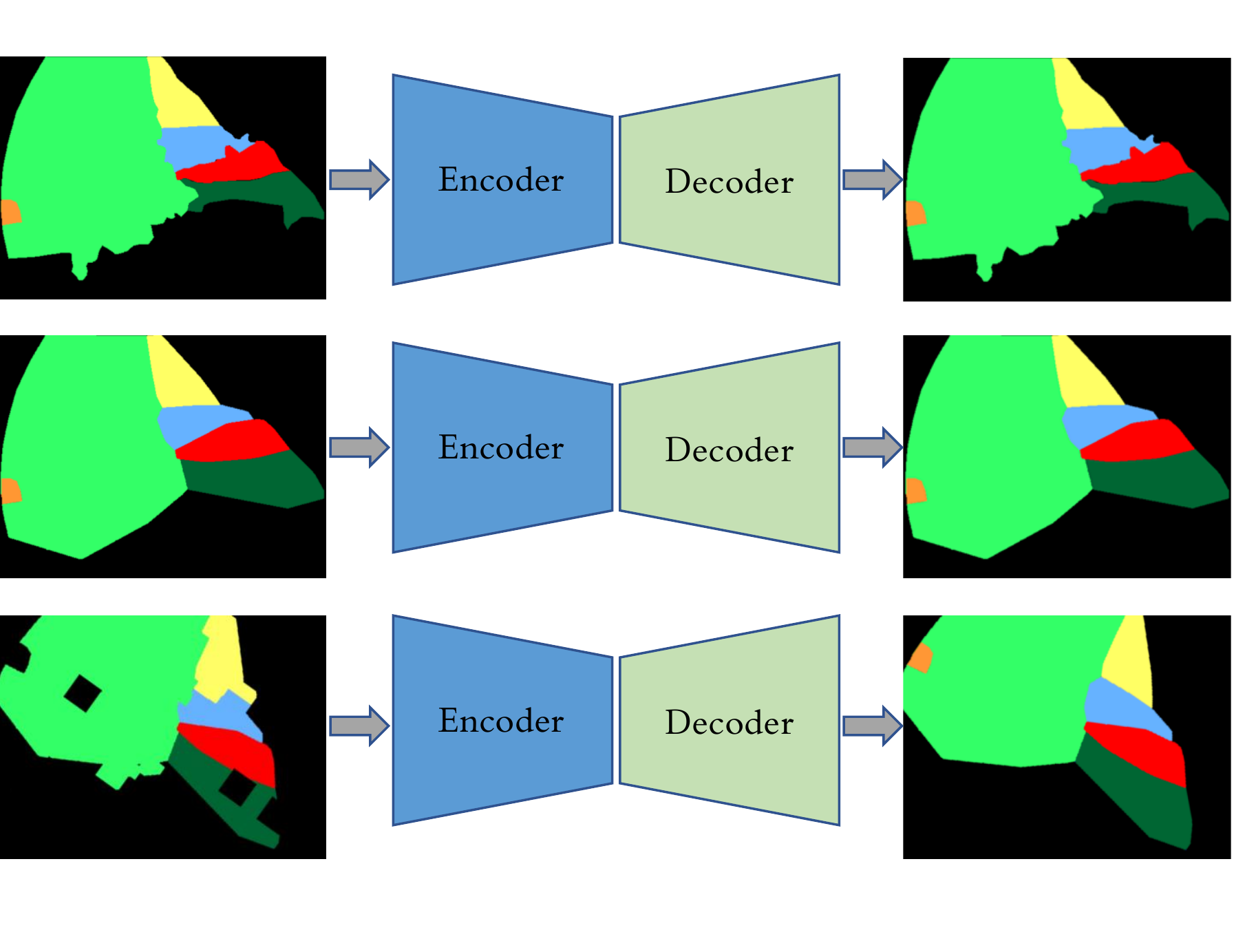}}
\caption{Augmentations strategies to train the autoencoder network.  The autoencoder to reconstruct ground truth masks (top row), augmented masks generated by overlaying convex envelopes of anatomical classes (middle row) and generated using affine transforms and morphological operations (bottom row).}
\label{figI}
\end{figure}

\begin{table*}[!t]
\centering
\caption{Class Distribution for Endoscapes.}
\begin{tabular}{cl>{\centering\arraybackslash}m{2.8cm}>{\centering\arraybackslash}m{2.8cm}>{\centering\arraybackslash}m{2.8cm}>{\centering\arraybackslash}m{2.8cm}}
\hline
\multirow{2}{*}{Category} & \multirow{2}{*}{Class Name} & \phantom{aaaaa}Endoscapes\phantom{aaaaa} (All Images)  & Training Set             & Validation Set           & Test Set                 \\ \cline{3-6} 
                          &                             & Instances per Class (\%) & Instances per Class (\%) & Instances per Class (\%) & Instances per Class (\%) \\ \hline
\multirow{6}{*}{Anatomy} & Background                   &       1933 (100\%)       &    1212 (100\%)          &       409 (100\%)         &     312 (100\%)         \\ \cline{2-6} 
                         & Cystic Plate                 &       701 (36.26\%)      &    433 (35.73\%)         &      138 (33.74\%)       &     130 (41.67\%)       \\ \cline{2-6} 
                          & HC Triangle (dissection)             &       683 (35.33\%)      &    426 (35.15\%)         &      133 (32.52\%)       &     124 (39.74\%)       \\ \cline{2-6} 
                          & Cystic Artery               &       1020 (52.77\%)     &    636 (52.48\%)         &      192 (46.94\%)       &     192 (61.54\%)        \\ \cline{2-6} 
                          & Cystic Duct                 &       1477 (76.41\%)     &    952 (78.55\%)         &      267 (65.28\%)       &     258 (82.69\%)        \\ \cline{2-6} 
                          & Gallbladder                 &       1812 (94.20\%)     &    1174 (96.86\%)        &       360 (88.01\%)      &      287 (91.99\%)       \\ \hline
\multirow{20}{*}{Instrument}   & Grasper Tip                 &       1386 (71.70\%)     &    879 (72.52\%)         &        277 (67.73\%)     &     230 (73.72\%)        \\ \cline{2-6} 
                          & Grasper Shaft               &       332 (17.18\%)      &    210 (17.33\%)         &        73 (17.85\%)      &     49 (15.70\%)         \\ \cline{2-6} 
                          & Bipolar Tip                 &       138 (7.14\%)       &    91 (7.51\%)           &        37 (9.05\%)       &     10 (3.20\%)          \\ \cline{2-6} 
                          & Bipolar Shaft               &       109 (5.64\%)       &    70 (5.78\%)           &        30 (7.34\%)       &     9 (2.88\%)           \\ \cline{2-6} 
                          & Hook Tip                    &       957 (49.51\%)      &    611 (50.41\%)         &        187 (45.72\%)     &     159 (50.96\%)        \\ \cline{2-6} 
                          & Hook Shaft                  &       825 (42.68\%)      &    536 (44.22\%)         &        156 (38.14\%)     &     133 (42.63\%)        \\ \cline{2-6} 
                          & Scissors Tip                &       5 (0.26\%)         &    3 (0.25\%)            &        0 (0\%)           &     2 (0.64\%)           \\ \cline{2-6} 
                          & Scissors Shaft              &       4 (0.21\%)         &    3 (0.25\%)            &        0 (0\%)           &     1 (0.32\%)           \\ \cline{2-6} 
                          & Clipper Tip                 &       211 (10.92\%)      &    120 (9.90\%)          &        48 (11.74\%)      &     43 (13.78\%)         \\ \cline{2-6} 
                          & Clipper Shaft               &       200 (10.35\%)      &    117 (9.65\%)          &        45 (11.00\%)      &     38 (12.18\%)         \\ \cline{2-6} 
                          & Plastic Clipper Tip         &       5 (0.26\%)         &    5 (0.41\%)            &        0 (0\%)           &     0 (0\%)              \\ \cline{2-6} 
                          & Plastic Clipper Shaft       &       1 (0.05\%)         &    1 (0.08\%)            &        0 (0\%)           &     0 (0\%)              \\ \cline{2-6} 
                          & Irrigator                   &       72 (3.72\%)        &    47 (3.88\%)           &        11 (2.69\%)       &     14 (4.49\%)          \\ \cline{2-6} 
                          & Retractor                   &       6 (0.31\%)         &    4 (0.33\%)            &        0 (0\%)           &     2 (0.64\%)           \\ \cline{2-6} 
                          & Bipolar Variant Tip         &       2 (0.10\%)         &    2 (0.16\%)            &        0 (0\%)           &     0 (0\%)              \\ \cline{2-6} 
                          & Bipolar Shaft Variant       &       1 (0.05\%)         &    1 (0.08\%)            &        0 (0\%)           &     0 (0\%)              \\ \cline{2-6} 
                          & Hook Variant Tip            &       101 (5.22\%)       &    71 (5.86\%)           &        15 (3.67\%)       &     15 (4.81\%)          \\ \cline{2-6} 
                          & Hook Shaft Variant          &       98 (5.07\%)        &    67 (5.53\%)           &        15 (3.67\%)       &     16 (5.13\%)          \\ \cline{2-6} 
                          & Clipper Variant Tip         &       1 (0.05\%)         &    1 (0.08\%)            &        0 (0\%)           &     0 (0\%)              \\ \cline{2-6} 
                          & Clipper Shaft Variant       &       1 (0.05\%)         &    1 (0.08\%)            &        0 (0\%)           &     0 (0\%)              \\ \hline
\multirow{3}{*}{Other}    & Lymph Node                   &       453 (23.44\%)      &    314  (25.91\%)        &        98 (23.96\%)      &     41 (13.14\%)         \\ \cline{2-6} 
                          & Undissected Gallbladder     &       12 (0.62\%)        &    6 (0.50\%)            &        6 (1.47\%)        &     0 (0\%)              \\ \cline{2-6} 
                          & Anatomical Variant          &       88 (4.55\%)        &    73 (6.02\%)           &        15 (3.67\%)       &     0 (0\%)              \\ \hline
\end{tabular}
\end{table*}

\subsection{Methodology}
Given a fully annotated semantic segmentation dataset $\mathbf{D} = \left\{ \left ( \mathbf{x },\mathbf{y}\right ); \: \mathbf{x}\in \mathbf{X} ,  \mathbf{y}\in \mathbf{Y}\right\}$, for each input image $\mathbf{x } = \left \{ x_{i};\: x_{i}\in \mathbb{R} , i \in \mathbb{S} \right\}$ a segmentation model ($\phi$) is trained to estimate the corresponding ground truth mask $\mathbf{y} = \left \{ y_{i};\: y_{i}\in \mathbb{C} , i \in \mathbb{S} \right\}$. Conventionally, convolutional neural networks (CNN) are trained to do this by minimizing a loss function $L_{seg}$ penalizing errors in predicting the ground truth mask. In most cases, $L_{seg}$ is chosen to supervise the model at a pixel level using functions such as cross entropy, dice loss, etc. However, these loss functions do not fully exploit global constraints and regularities in the output space. As described in \cite{acnn}, this approach may be suboptimal for datasets depicting objects that are highly regular in their appearance or position such as in the case of anatomical structures in medical images. In \cite{acnn}, the authors make use of an autoencoder network to penalize global features in addition to a cross entropy loss term. To do this, they first train an autoencoder network, comprising an encoder and decoder, to be able to reconstruct segmentation masks. Internally, the encoder ($\psi$) learns a hidden representation, $\psi(\mathbf{y})$, which the decoder uses to reconstruct the original mask. If the hidden representation is undercomplete, i.e. of lower dimensionality than the input mask, the autoencoder learns to represent only the most salient, or global, features of the input mask in the encoding while discarding more granular, or local, features. They then balance their local cross entropy loss function with a loss term penalizing more global errors in the prediction:

\begin{equation}
L_{global} = d(\psi(\phi(\mathbf{x})), \; \psi(\mathbf{y})).
\end{equation}

Here, $d$ is a metric function quantifying the distance between the encodings for the predicted and ground truth mask. This loss term ensures that the prediction is close to the ground truth mask in the same low dimensional space, for example on a shape manifold. We further hypothesize that if such a low dimensional space can be learned, then the predicted masks should move continuously across that space over time. We incorporate this into the training scheme in order to provide temporal supervision using an additional contrastive loss term. For three temporally consecutive predicted masks $\phi (\mathbf{x_{t}} )$(anchor), $\phi (\mathbf{x_{t+1}} )$(positive), and $\phi (\mathbf{x_{t+2}} )$(negative), we use a triplet loss constraining the three corresponding encodings $e_a$, $e_p$, and $e_n$, respectively. 

\begin{equation}
L_{consistency} = max(0, d(e_{a}, e_{p}) - d(e_{a}, e_{n})).
\end{equation}
Using the additional unlabeled frames, this term penalizes abrupt changes in the encodings of predicted masks at consecutive timesteps, and consequently, abrupt changes in global characteristics of predicted masks over time. In this semi-supervised setup, we then balance the standard pixel-wise loss functions with the consistency loss and global loss as described below
\begin{equation}
Loss = L_{seg}+ \lambda _{g}L_{global}+ \lambda _{c}L_{consistency}.
\end{equation}

We would like to emphasize that this framework is both model-agnostic and incurs no additional computational cost at inference time.

\subsection{Training Strategies}
Our proposed approach is trained in two stages. The aim of the first stage is to learn a lower-dimensional parameterization of a predicted or ground truth segmentation mask that captures the structure and configuration of the represented classes \cite{acnn}\cite{vconv-dae}. To this end, we train an autoencoder network to generate an undercomplete representation, i.e. of lower dimensionality than the input, by learning to reconstruct ground truth segmentation masks. This stage of training is crucial to learning a compact and expressive representation space that can effectively supervise the segmentation network training. The fact that segmentation masks are significantly less detailed and more regular in appearance than surgical images allows for new and more robust augmentation strategies to support this stage of training. Using this property, we propose a number of strategies to simulate segmentation masks and inflate the dataset size: (1) We generate plausible artificial masks by overlaying the convex closure of each anatomical class in a ground truth segmentation mask. We do not modify the tool classes as these correspond to non-deformable classes with characteristic shapes; (2) We perform affine transforms such as rotation and rescaling; (3)  Following \cite{postdae1}, we use erosion and dilation with variable kernels to degrade the input in order to force the network to ignore minor local variations in the input. Empirically, we observe that these additional augmentation strategies and a deeper CNN design choice for the autoencoder were critical to getting reliable reconstruction accuracy, particularly on highly dynamic data such as laparoscopic videos. A brief overview of the first stage of training and a depiction of the augmentation strategies used is shown in Fig. 2.

In the second stage, we train a video segmentation model in a semi-supervised manner and use the encoding phase of the trained autoencoder to provide additional supervision as described previously. We freeze the autoencoder weights during this phase of training to avoid the network collapsing to a trivial solution to satisfy the encoding space constraints.

\begin{table}[]
\centering
\caption{Autoencoder Structure: Under the column Layer: Conv - Convolutional Layer, ConvT - Transpose Convolutional layer, FC - Fully Connected layer. All convolutional and transpose convolutional layers use a 3x3 kernel with number of filters, stride and activation described by the Num Outputs, Stride and Activation columns, respectively. The second FC layer regenerates the same number of outputs as the number of features entering the first FC layer, which is dependent on input size.}
\scalebox{1.25}{
\begin{tabular}{l>{\centering\arraybackslash}m{1.3cm}>{\centering\arraybackslash}m{1.3cm}>{\centering\arraybackslash}m{1.3cm}}
\hline
Layer & Num Outputs & Stride & Activation \\ \hline
Conv  & 16          & 2      & ReLU       \\
Conv  & 16          & 1      & ReLU       \\
Conv  & 32          & 2      & ReLU       \\
Conv  & 32          & 1      & ReLU       \\
Conv  & 64          & 2      & ReLU       \\
Conv  & 64          & 1      & ReLU       \\
Conv  & 64          & 2      & ReLU       \\
Conv  & 64          & 1      & ReLU       \\
FC    & 1024        & -      & None       \\ \hline
FC    & -           & -      & None       \\
ConvT & 64          & 2      & ReLU       \\
Conv  & 64          & 1      & ReLU       \\
ConvT & 64          & 2      & ReLU       \\
Conv  & 64          & 1      & ReLU       \\
ConvT & 64          & 2      & ReLU       \\
Conv  & 64          & 1      & ReLU       \\
Conv  & 32          & 2      & ReLU       \\
ConvT & 32          & 1      & ReLU       \\
Conv  & \# classes & 1      & None   \\\hline
\end{tabular}
}
\end{table}

\begin{table}[!]
\caption{Results Endoscapes: From top to bottom, the F1 scores are presented per class followed by the mean F1, IoU and Pixel Accuracy over all considered classes. The per class results are ordered from the least represented class in the dataset to the most represented.}
\scalebox{1.0}{
\begin{tabular}{l>{\centering\arraybackslash}m{1.3cm}>{\centering\arraybackslash}m{1.3cm}>{\centering\arraybackslash}m{1.3cm}}
\hline
\backslashbox{Class Name}{Model}& Image Predictor & Video Predictor & Ours \\ \hline
F1 HC Triangle                & 60.72 \%        & 62.63 \%                & \textbf{64.39 \%}                       \\ \hline
F1 Cystic Plate                & 39.71 \%        & 41.74 \%                & \textbf{48.34 \%}                       \\ \hline
F1 Cystic Artery               & 47.74 \%        & 46.31 \%                & \textbf{50.79 \%}                       \\ \hline
F1 Cystic Duct                 & 61.06 \%        & 62.06 \%                & \textbf{62.55 \%}                       \\ \hline
F1 Instruments                 & 94.87 \%        & 94.41 \%                & \textbf{94.90 \%}                       \\ \hline
F1 Gallbladder                 & 92.16 \%        & 92.06 \%                & \textbf{92.99 \%}                       \\ \hline
F1 Background                  & 97.22 \%        & 97.16 \%                & \textbf{97.43 \%}                       \\ \hline
Mean F1                        & 70.50 \%        & 70.91 \%                & \textbf{73.06 \%}                       \\ \hline
Mean IoU                       & 59.14 \%        & 59.47 \%                & \textbf{61.58 \%}                       \\ \hline
Mean Pixel Accuracy            & 94.78 \%        & 94.73 \%                & \textbf{95.22 \%}                       \\ \hline
\end{tabular}
}
\end{table}

\begin{table}[!]
\caption{Results CaDIS: From top to bottom, the F1 scores are presented per class followed by the mean F1, IoU and Pixel Accuracy over all considered classes. The per class results are ordered from the least represented class in the dataset to the most represented.}
\scalebox{1.0}{
\begin{tabular}{l>{\centering\arraybackslash}m{1.3cm}>{\centering\arraybackslash}m{1.3cm}>{\centering\arraybackslash}m{1.3cm}}
\hline
\backslashbox{Class Name}{Model} & Image Predictor & Video Predictor & Ours \\ \hline
F1 Hand                         & 94.59 \%    & 92.73 \%                & \textbf{95.29 \%}                       \\ \hline
F1 Eye Retractors               & 80.31 \%    & 82.40 \%                & \textbf{83.58 \%}                       \\ \hline
F1 Surgical Tape                & 88.82 \%    & 89.52 \%                & \textbf{90.06 \%}                       \\ \hline
F1 Instrument                   & 79.68 \%    & 81.72 \%                & \textbf{83.22 \%}                       \\ \hline
F1 Pupil                        & 96.51 \%    & 96.26 \%                & \textbf{96.55 \%}                       \\ \hline
F1 Skin                         & 86.75 \%    & \textbf{90.81 \%}       & 90.77 \%                       \\ \hline
F1 Iris                         & 89.80 \%    & 89.98 \%                & \textbf{90.22 \%}                       \\ \hline
F1 Cornea                       & 93.58 \%    & 95.17 \%                & \textbf{95.21 \%}                       \\ \hline
Mean F1                        & 88.76 \%     & 89.82 \%                & \textbf{90.61 \%}                       \\ \hline
Mean IoU                       & 80.28 \%     & 81.90 \%                & \textbf{83.18 \%}                       \\ \hline
Mean Pixel Accuracy            & 91.62 \%     & 93.11 \%                & \textbf{93.30 \%}                       \\ \hline
\end{tabular}
}
\end{table}

\begin{figure*}[]
\centerline{\includegraphics[clip,trim=0.0cm 4cm 0cm 7.6cm, width=18cm]{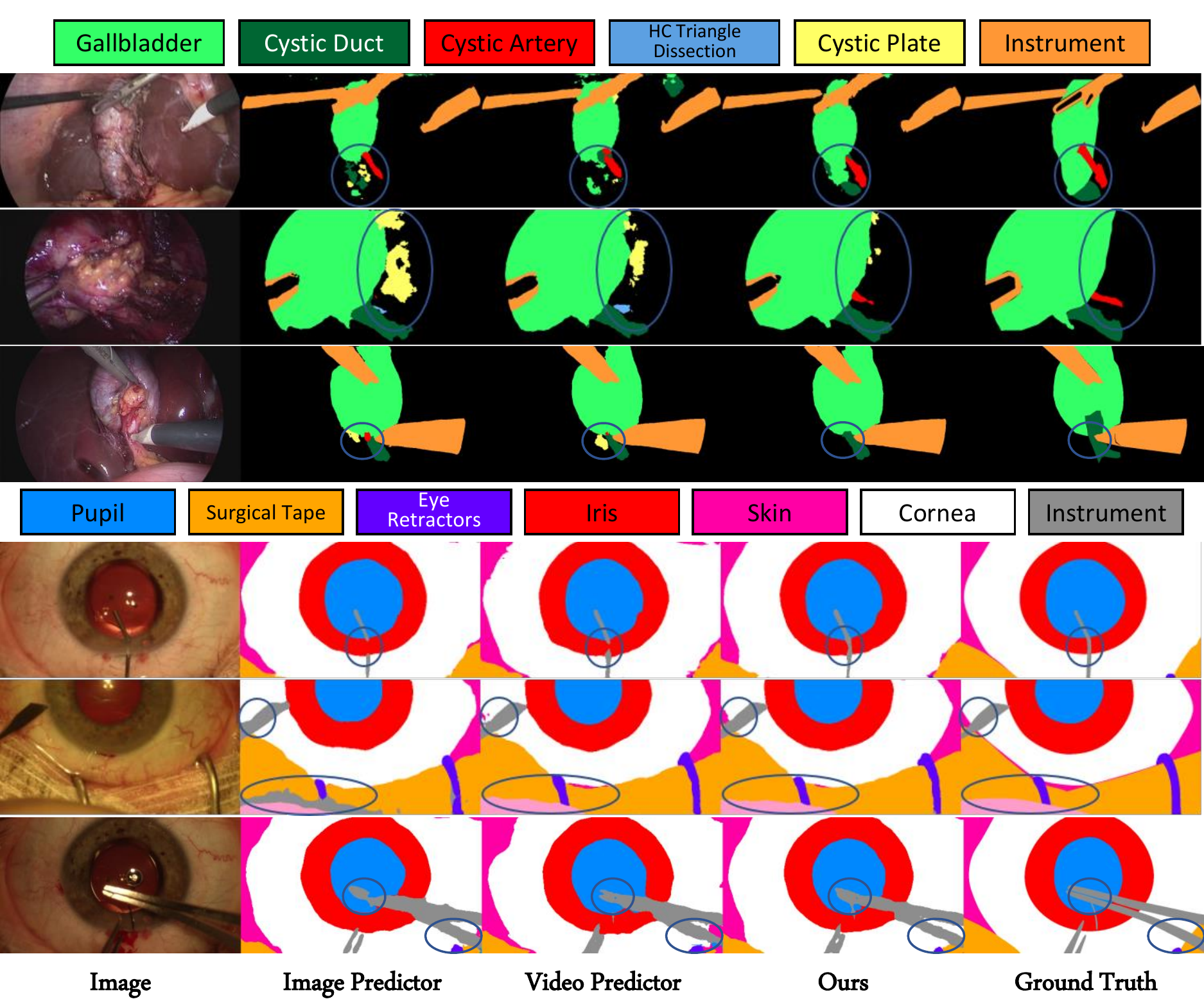}}
\caption{Qualitative Results: The additional supervision incorporated into the training leads to a significant decrease in disconnected blobs, irregular boundaries and missing classes as highlighted in the circled regions.}
\label{fig3}
\end{figure*}

\section{Experimental Setup}

We benchmark our algorithm on two surgical video datasets depicting sequences from laparoscopic cholecystectomies and cataract surgeries. We establish two baselines, Deeplab-v3+, a state-of-the-art method for image semantic segmentation used in this setting in \cite{mascagni2021artificial}, and an adapted version of the same model designed to propagate temporal information to make temporally coherent predictions. Hereafter, we refer to these models as Image Predictor and Video Predictor, respectively. On both datasets, we analyze the results per class and observe a consistent improvement over the baseline methods in performance when training using the proposed approach.
\label{sec:guidelines}

\begin{table*}[!]
\caption{Ablation Study.}
\centering
\scalebox{1.2}{
\begin{tabular}{l|>{\centering\arraybackslash}m{1.3cm}>{\centering\arraybackslash}m{1.3cm}>{\centering\arraybackslash}m{1.3cm}|>{\centering\arraybackslash}m{1.7cm}>{\centering\arraybackslash}m{1.7cm}>{\centering\arraybackslash}m{1.7cm}}
\cline{1-7}
Dataset                     & Temporal Module & Global loss & Consistency loss & mean IoU  & mean F1  & mean Pixel Accuracy  \\ \cline{1-7}
\multirow{5}{*}{Endoscapes} & \xmark          & \xmark      & \xmark           & 59.14 \%  & 70.5 \%  & 94.78 \%             \\ \cline{5-7}
                            & \xmark          & \checkmark  & \xmark           & 59.25 \%  & 70.42 \% & 95.04 \%             \\ \cline{5-7}
                            & \checkmark      & \xmark      & \xmark           & 59.47 \%  & 70.91 \% & 94.73 \%             \\ \cline{5-7}
                            & \checkmark      & \xmark      & \checkmark       & 60.5 \%   & 71.96 \% & 94.94 \%             \\ \cline{5-7}
                            & \checkmark      & \checkmark  & \xmark           & 60.92 \%  & 72.39 \% & 95.09 \%             \\ \cline{2-7}
                            & \checkmark      & \checkmark  & \checkmark       & \textbf{61.58 \%}  & \textbf{73.06 \%} & \textbf{95.22 \%}             \\ \cline{1-7}
\multirow{5}{*}{CaDIS}      & \xmark          & \xmark      & \xmark           & 80.28 \%  & 88.76 \% & 91.62 \%             \\ \cline{5-7}
                            & \xmark          & \checkmark  & \xmark           & 79.68 \%  & 88.45 \% & 91.22 \%             \\ \cline{5-7}
                            & \checkmark      & \xmark      & \xmark           & 81.9 \%   & 89.82 \% & 93.11 \%             \\ \cline{5-7}
                            & \checkmark      & \xmark      & \checkmark       & 82.35 \%  & 90.12 \% & 92.9 \%              \\ \cline{5-7}
                            & \checkmark      & \checkmark  & \xmark           & 82.37 \%  & 90.14 \% & 92.74 \%             \\ \cline{2-7}
                            & \checkmark      & \checkmark  & \checkmark       & \textbf{83.18 \%}  & \textbf{90.61 \%} & \textbf{93.3 \%}              \\ \cline{1-7}
\end{tabular}
}
\end{table*}

\subsection{Study Design}
To design the Video Predictor, we adapt Deeplabv3+\cite{hq1} to propagate temporal information using a 3x3 ConvLSTM \cite{clstm} module placed between the encoder and  decoder. The base architecture for Deeplabv3+ is a modified resnet-50\cite{resnet} initialized with Imagenet pretrained weights. The autoencoder is a 18 layer convolutional neural network. Given the design choice for the Video and Image Predictor, this corresponds to $\sim$5\% additional network capacity during training. To note, in the inference phase, the autoencoder network can be discarded and inference is performed without any additional complexity. The autoencoder network architecture is detailed in Table 2.

To train each model, we feed a temporally regularly spaced sequence of 3 unlabeled frames and the following labeled frame. To find the optimal weighting for the loss terms for each dataset, we explore combinations of $\lambda_{c}$ in \{0.0001, 0.001, 0.01, 0.1\} and $\lambda_{g}$ in \{0.5, 1, 1.5, 2\} using L2 and mean square distance as the distance function for $L_{consistency}$ and  $L_{global}$, respectively.

\subsection{Datasets}
To evaluate the model in a variety of surgical scenarios, we test the TCNN framework on two surgical video datasets with greatly differing appearance and motion patterns, Endoscapes and CaDIS. The former comprises recordings of laparoscopic cholecystectomy procedures and the latter recordings of microscopic cataract surgeries. Since cataract procedures are performed using a mounted microscope focused on the patient's eye, the appearance of the scene is much more stable in terms of scale, orientation and motion during and between procedures compared to laparoscopic surgeries.

\subsubsection{Endoscapes}
The Endoscapes dataset introduced in Section 3 was distributed as follows: 120 videos, 41 videos, and 40 for training, validation, and testing, respectively. A detailed description of the dataset splits is presented in Table 1.

To mitigate the label imbalance issue, which is not the primary focus of this work, we take a number of steps. 
\begin{itemize}
    \item We ignore all classes that appear in under 5\% of the images in any of the training, validation, or test splits during both the training and evaluation phase.
    \item We group all the remaining instrument tips and shafts into a single class.
    \item We ignore the lymph node class due to its highly variable appearance, position, and ambiguous boundaries.
\end{itemize}
We sample frames of resolution 854x480 at 1 frame per second (fps) to generate training sequences.

\subsubsection{CaDIS}
CaDIS\cite{cadis} is an image semantic segmentation dataset containing fully annotated labels for 25 videos or 4671 images. Of this, 3550 frames (19 videos) were used for training, 534 frames (3 videos) for validation, and 587 frames (3 videos) for testing following\cite{cadis}. This dataset was generated by labeling a subset of frames from the training videos of CATARCTS\cite{cataracts} dataset. In order to generate temporal sequences, we merge preceding frames available in the CATARACTS dataset with labels from CaDIS.
 
We follow the first experimental setup in \cite{cadis} to select class labels and downsize the images to a resolution of 270x480 during both training and evaluation. In this setup, 4 classes correspond to anatomical structures, 1 to instruments and 3 to miscellaneous structures. We sample frames at 15 fps in order to generate training sequences.

\subsection{Evaluation Metrics}

Dice Similarity coefficient (F-score), Intersection over Union (IoU), and Pixel Accuracy were used to evaluate the models. Pixels belonging to an ignored class were excluded from evaluation. 

\subsection{Implementation Details}

All models were trained on a single NVIDIA V100 for 200 epochs using a momentum optimizer with momentum=0.9, using the polynomial learning rate schedule described in \cite{hq1} with the initial and final learning rates set as 7e-3 and 1-6, respectively. The best performing model was selected during training based on mIoU on the validation set. To train the segmentation network, we set $\lambda_{c}=0.01$ for Endsocapes, $\lambda_{c}=0.001$ for CaDIS and $\lambda_{g}=1.5$  and in all presented results based on performance on the validation set.

\begin{figure*}[]
\centerline{\includegraphics[clip,trim=0.0cm 4cm 0cm 1cm, width=18cm]{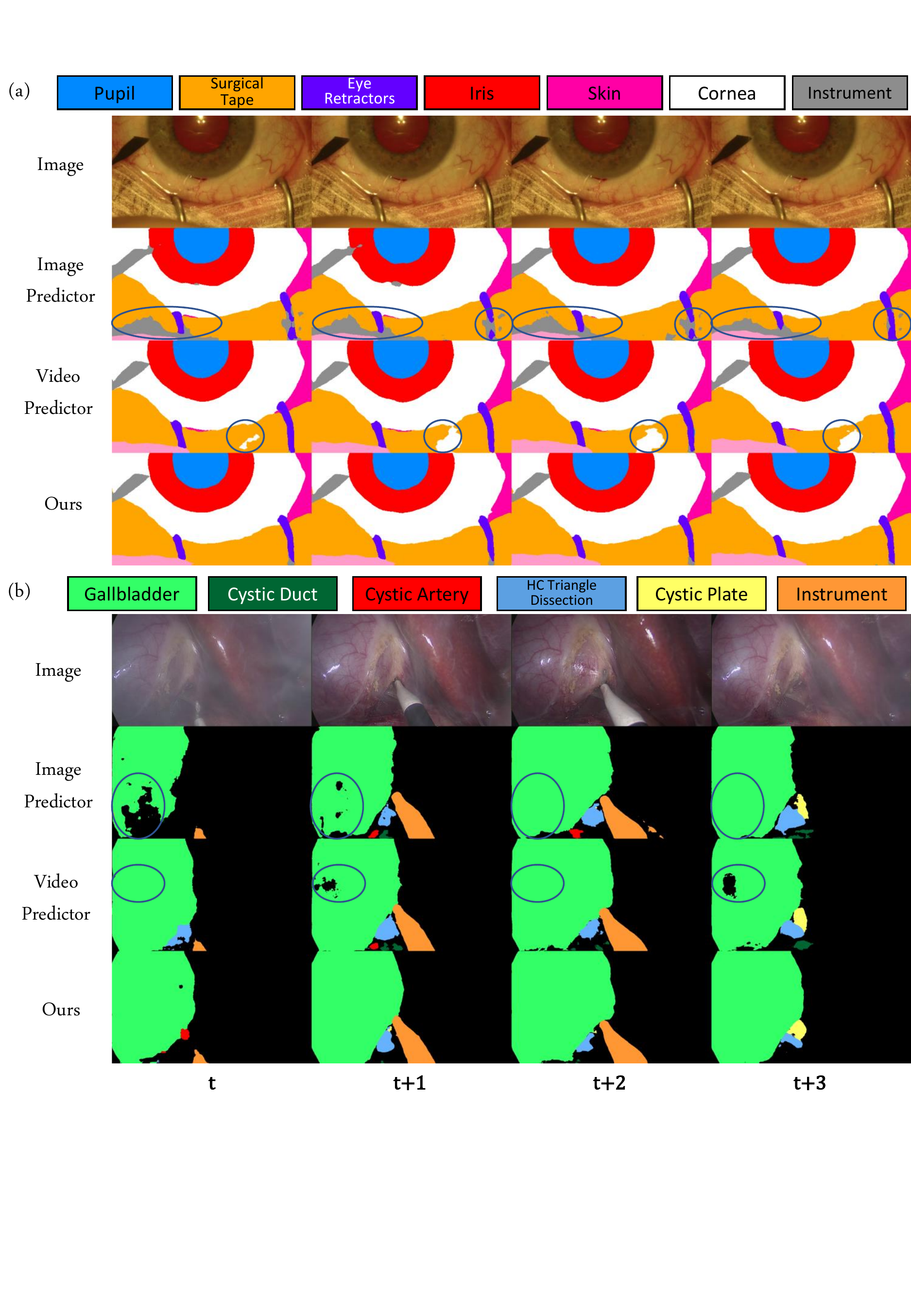}}
\caption{Example results showing the progression of predictions over consecutive timesteps. Case (a) depicts a sequence from the CaDIS test set showing how the addition of the temporal module in the Video Predictor allows for a more consistent prediction over time and further how the additional supervision leads to a more logical one. Case (b) shows a sequence from the Endoscapes test set showing how the TCNN framework makes the model less susceptible to error when dealing with smoke, a common confounding during surgical procedures. }
\label{fig4}
\end{figure*}

\section{Results}
In Table 3 and Table 4, we present comparative results for the Image Predictor, Video Predictor, and our approach using the TCNN framework. Specifically, we observe an average improvement of 2.6\% and 1.9\%  F1 over the Image Predictor on Endoscapes and CaDIS, respectively. Of note, while there is an improvement in the average performance over all classes, similar to \cite{pfeuffer} \cite{exploitingtemp}, not all classes benefit from the addition of the temporal module by itself. This may be because the increased network capacity leads the model to overfit to the training data, indicating that additional supervision could be necessary. This is reinforced by the fact that there is a marked and consistent improvement when using the global and consistency losses.  On both datasets, we observe that the least represented classes benefit the most from the additional supervision. Using the additional loss terms, we obtain a mean improvement of 4.3\% and 1.4\% over the Video Predictor on the 3 least represented classes in Endoscapes and CaDIS, respectively, at the exact same computational cost. On the Endoscapes dataset, there is a substantial performance improvement in the identification of the Cystic Plate class with a gain of 6.6\% F1 over the Video Predictor. While this class does not have a characteristic appearance because it is defined surgically by the quality of the dissection, it does have more particular global characteristics, such as position, that may be better captured using our proposed framework. We provide some qualitative results in Fig. 3 to illustrate the benefits of the additional supervision. We see that the TCNN framework helps minimize the occurrence of improbable predictions such as isolated blobs and anatomical structures at unrealistic positions.  In Fig. 4, we further demonstrate how the additional loss terms help make more consistent predictions over consecutive timesteps, making the network more robust to confounding factors such as smoke and occlusions.

Finally, we present the results of the ablation study in Table 5. Interestingly, the global loss term proves more effective when used in conjunction with the temporal module. Specifically, global loss contributes to an increase of 1.48\% and 0.32\% F1 for the Video Predictor on Endoscapes and CaDIS, respectively. In contrast, it results in a decrease in performance of the image predictor by 0.08\% and 0.31\% F1 on the two datasets, respectively. Note, that this case corresponds to the method proposed in \cite{acnn}. We hypothesize that the additional temporal context provided through the temporal module allows the network to better rectify global discrepancies in the predicted mask. Finally, on both datasets we observe that the consistency loss provides a consistent improvement and works well in combination with the global loss.


\section{Conclusion}
In this work, we present a framework, TCNN, for semi-supervised video semantic segmentation utilizing autoencoder networks to provide spatio-temporal regularization for convolutional neural networks. Our approach is easy to implement and is independent of the model design, providing additional supervisory signals to arbitrary video semantic segmentation models. With minimal additional computational cost during training, the TCNN framework shows consistent improvement in segmentation performance on multiple datasets with exactly the same runtime and computational cost during inference. We hope the work on this framework and presented datasets will enhance video semantic segmentation in settings with limited data and intrinsic topological constraints and promote
research into surgical video segmentation.
\\\\
{\bfseries Ethical approval} The surgical videos were recorded and collected in an anonymized manner following the informed consent of patients in compliance with the local Institutional Review Board (IRB) requirements.

\bibliography{citations}

\end{document}